\documentclass[sigconf]{acmart}
\AtBeginDocument{%
  \providecommand\BibTeX{{%
    \normalfont B\kern-0.5em{\scshape i\kern-0.25em b}\kern-0.8em\TeX}}}

\setcopyright{acmcopyright}
\copyrightyear{2018}
\acmYear{2018}
\acmDOI{10.1145/1122445.1122456}
\usepackage{multirow}
\usepackage{bm}

\usepackage[belowskip=5pt,aboveskip=0pt]{caption}
\usepackage{mathrsfs}
\usepackage{xcolor}
\acmConference[Woodstock '18]{Woodstock '18: ACM Symposium on Neural
  Gaze Detection}{June 03--05, 2018}{Woodstock, NY}
\acmBooktitle{Woodstock '18: ACM Symposium on Neural Gaze Detection,
  June 03--05, 2018, Woodstock, NY}
\acmPrice{15.00}
\acmISBN{978-1-4503-XXXX-X/18/06}

\begin{document}

\title{Online Misinformation Detection in Live Streaming Videos}


\author{Rui Cao}
\email{ruicao.2020@phdcs.smu.edu.sg}
\affiliation{%
  \institution{Singapore Management University}
  \country{Singapore}
  \city{Singapore}
 }
\renewcommand{\shortauthors}{Rui Cao et al.}


\begin{abstract}
Online misinformation detection is an important issue and methods are proposed to detect and curb misinformation in various forms. 
However, previous studies are conducted in an offline manner. We claim a realistic misinformation detection setting that has not been studied yet is online misinformation detection in live streaming videos (MDLS).
In the proposal, we formulate the problem of MDLS and illustrate the importance and the challenge of the task. Besides, we propose feasible ways of developing the problem into AI challenges as well as potential solutions to the problem.
\end{abstract}




\maketitle

\section{Problem Statement}
\label{sec:intro}
The development of online social platforms facilitates people in idea sharing and information exchange. However, these platforms are increasingly exploited for the propagation of misinformation~\cite{DBLP:journals/peerj-cs/DUliziaCFG21,DBLP:journals/csur/ZhouZ20}. 
Several methods have been proposed to combat the spread of misinformation in various settings, from uni-modal texts~\cite{DBLP:conf/coling/Perez-RosasKLM18,DBLP:conf/acl/WongGM18} to multi-modal memes~\cite{DBLP:conf/tto/JabiyevOSK21} and videos~\cite{ganti2022novel,DBLP:journals/corr/abs-2107-00941}, whereas, in an offline manner.
With the prevalence of live streaming, we claim that it is urgent to detect misinformation in a real-time online manner. A realistic misinformation detection setting that has not been well studied is misinformation detection in live streaming videos (\textbf{MDLS}).

There are two major differences between misinformation detection in offline videos and MDLS. Firstly, offline misinformation detection has access to complete videos while MDLS continuously receives and processes new video frames (optionally associated with audio segments) but without any information from future. Secondly, to combat the spread of misinformation via living streaming, the detection of misinformation should be as early as possible once misinformation contents appear.

We formally define the task of MDLS as follows by referring to the testing pipeline. At the $t$-th time step during the inference period, the system receives a frame $\mathcal{F}_t = (\mathcal{I}_t, \mathcal{A}_t)$ from the live streaming video, where $\mathcal{I}_t$ is the image for the frame and $\mathcal{A}_t$ is the associated audio. The system is required to predict whether the $t$-th time step involves misinformation given $(\mathcal{F}_1, \mathcal{F}_2, \cdots, \mathcal{F}_t)$. For a living streaming video, if all time steps are not predicted as misinformation, the video will be regarded as neutral, otherwise, misinformation. A good system should detect misinformation live streaming videos among neutral ones and detect as soon as possible when frames containing misinformation are shown.
We provide more details about data annotation in Section~\ref{sec:dataset} and details for model evaluation in Section~\ref{sec:evaluation}.

\section{Value Proposition}
\label{sec:value}
The misinformation dissemination has evolved from uni-modal based propagation~\cite{DBLP:conf/icip/HeLW19,DBLP:journals/corr/ZubiagaHLPT15} to a more expressive manner, which integrates information from multiple modalities~\cite{DBLP:conf/tto/JabiyevOSK21,HUANG2022102967}. The misinformation detection under a multi-modal setting is even more challenging as it asks for comprehension of each modality as well as interaction between modalities. Several works tried to address multi-modal misinformation detection~\cite{DBLP:conf/tto/JabiyevOSK21,ganti2022novel,DBLP:journals/s00530-022-00966-y}. However, to the best of our knowledge, current study are related to offline detection and the problem of detecting live streaming misinformation videos has not been studied yet. 

Nowadays, it is possible for everyone to be a live streamer with the supports from social platforms, like TikTok and YouTube. With few constraints, malicious users can easily create and post misinformation via live streaming\footnote{https://www.business-standard.com/article/international/tiktok-suspends-live-streaming-in-russia-citing-new-fake-news-law-122030700021\_1.html, 
https://www.engadget.com/twitch-misinformation-policy-vaccines-russia-192523783.html}. On the other hand, live streaming plays an important role in propagation of information in different aspects, such as live broadcast of sport games, TV news and E-commerce, so that misinformation in live streaming videos has impact to a great number of people. With the rapidly growing number of diverse live streaming contents, it is infeasible for platforms to detect misinformation manually. Therefore, it is an urgent topic to design automatic methods to detect misinformation in live streaming videos. However,
the detection of online misinformation live streaming videos has not been explored yet.

Our proposed challenge can be said to serve the dual purpose of broadening the research scope of misinformation detection in an online manner and at the same time facilitating the real-world application of live streaming to curb misinformation.

\section{Existing Solutions}
\label{sec:related}
Automatic detection of online misinformation has received great attention from the community for smart city sustainability. As fake messages usually involve fake facts, 
various automatic fact-checking methods are proposed, which check related knowledge bases and compare the pieces of knowledge appearing in posts~\cite{DBLP:journals/corr/CiampagliaSRBMF15,DBLP:journals/isci/PasiVC19}. Besides direct check about the validity of information, some studies consider detecting misinformation in terms of specific styles of misinformation~\cite{DBLP:journals/corr/abs-1904-11679,DBLP:conf/coling/Perez-RosasKLM18} or ways of propagation~\cite{DBLP:conf/icde/WuYZ15,DBLP:conf/acl/WongGM18}.
Meanwhile, the scope of misinformation detection has broaden from uni-modal texts~\cite{DBLP:journals/ipm/BarbadoAI19,DBLP:journals/corr/ZubiagaHLPT15} or images~\cite{DBLP:conf/icip/HeLW19,DBLP:journals/tifs/GuoCZW18}, to multi-modal contents~\cite{HUANG2022102967,DBLP:journals/pacmhci/HusseinJM20,DBLP:conf/tto/JabiyevOSK21}.
However, to the best of our knowledge, we are the first study exploring the problem of detecting misinformation live streaming videos.

The most similar work to ours is misinformation detection in offline videos~\cite{HUANG2022102967,ganti2022novel}. In~\cite{HUANG2022102967}, videos are converted into captions. The textual captions are processed into vector representations for classification. It transforms the multi-modal task into a uni-modal classification task. The proposed model in~\cite{ganti2022novel} is three-stage, paying attention to both visual and textual information of videos step-by-step.
The major difference between misinformation detection in offline videos and live streaming videos is the later one receives partial inputs and requires timeliness. 

One of the most similar task requiring timeliness is action detection in live streaming videos~\cite{DBLP:conf/eccv/ShouPCMMVNC18,DBLP:conf/iccv/GaoX0SX19,DBLP:journals/ijcv/NguyenT14}. The task requires detection of actions in live streaming video as early as possible, leading to partial observation of videos. The task can be treated as a frame-level binary classification task, where early starts of actions are treated as positive examples while the rests are negative ones. The max-margin model in~\cite{DBLP:journals/ijcv/NguyenT14} is based on Structured Output SVM (SOSVM)~\cite{tsochantaridis2005large} and extends SOSVM to sequential data. It aims to learn a detector that correctly classifies partial actions. The following work~\cite{DBLP:conf/eccv/ShouPCMMVNC18} considers longer and more complicated videos, where the type of actions are not known beforehand and multiple actions may appear in a video. It trains a classifier which outputs lower scores for backgrounds and  a higher score immediately an action starts and detects the class of the action simultaneously. A more advanced model~\cite{DBLP:conf/iccv/GaoX0SX19} separates the detection and classification of actions in two steps.
Our setting is mostly similar to~\cite{DBLP:journals/ijcv/NguyenT14}, while there will be multiple misinformation periods (positive segments). 

\section{Dataset Construction}
\label{sec:dataset}

\subsection{Definition}
\label{sec:data-define}
\noindent\textbf{Misinformation Video}
Following previous work~\cite{guess2019less,DBLP:journals/corr/abs-2107-00941}, we define ``misinformation'' as \textit{incorrect} or \textit{misleading information} that deceives users. We consider a video \textit{neutral} (not misinformation) if the whole video does not contain any misinformation contents otherwise, misinformation.

\noindent\textbf{Misinformation Live Streaming Videos}
We assume a video consists of a sequence of frames with associated audio $\mathcal{F} = \{\mathcal{F}\}_{t=1}^{T}$, where $T$ is the length of the video, $\mathcal{F}_t= (\mathcal{I}_t,\mathcal{A}_t)$, and $\mathcal{I}_t$ and $\mathcal{A}_t$ are the image and audio for the $t$-th frame respectively. We regard a live streaming video to be misinformation if frames in a video segment $\mathcal{S}^k=(\mathcal{F}_{\text{s}}^k,\mathcal{F}_{\text{s}+1}^k,\cdots,\mathcal{F}_{\text{e}}^k)$ are related to misinformation, where $k$ represents the $k$-th misinformation segment in the video and $\text{s}$ and $\text{e}$ represent the start and end of the segment.

\subsection{Video Collection}
We provide two alternative approaches for collecting videos, by either re-using previous collected and annotated data or collecting data ourselves.

\noindent\textbf{Re-use Data:} In~\cite{DBLP:journals/pacmhci/HusseinJM20}, a misinformation video dataset was proposed covering five topics. Videos are annotated into three categories: \textit{neutral}, \textit{misinformation} and  \textit{debunking misinformation}, 
where each unique video can be segmented into multiple training and testing instances~\cite{DBLP:journals/pacmhci/HusseinJM20}.

\noindent\textbf{Collect Data:} Alternatively, we can collect and annotate videos in a similar manner to~\cite{DBLP:journals/pacmhci/HusseinJM20}. Firstly, we choose topics which are frequently related to misinformation. Based on topics,  we collect videos from online platforms (i.e., YouTube and TikTok). For video annotation, we follow the annotation heuristics in~\cite{DBLP:journals/pacmhci/HusseinJM20} and elaborate details in Section~\ref{sec:anno-process}.

\subsection{Annotation Process}
\label{sec:anno-process}
Besides annotating whether the whole video is misinformation or not, in the live streaming setting, we are also asking starts of misinformation spans in the video. We define two levels of annotation: video-level and frame-level annotation. 

\noindent\textbf{Video-level Annotation:} 
The video-level annotation is the same as the annotation for misinformation videos.
Given a video, we ask annotators to classify it into either \textit{neutral} or  \textit{misinformation} given the definition in Section~\ref{sec:data-define}. Annotators are trained in advance with examples of misinformation videos to make sure they understand the definition. We can also conduct a dry run and compute the inter-annotator agreement in terms of Cohen’s $\kappa$~\cite{DBLP:conf/ranlp/BobicevS17}. Increment of $\kappa$ is expected after training annotators. 
Then, each video will be given to three annotators. We use the majority voting to decide the final label of each video. 

\noindent\textbf{Frame-level Annotation:}
For videos that are annotated as \textit{misinformation}, we ask annotators to point out starts and ends of spans which are related to misinformation contents. Similar to the video-level annotation, we will give annotation instructions. For consistency, the annotation will be made by one annotator in two passes. If annotations in two passes vary much, we will give a warning to the annotator and ask for the third annotation. The result will be verified by another annotator to ensure the detected spans are related to misinformation. The annotation can be done via Viper~\footnote{http://viper-toolkit.sourceforge.net/products/gt/}.

\section{Evaluation Metrics}
\label{sec:evaluation}
In the online live streaming misinformation video detection, the objective is more than the general goal of offline misinformation detection, which requires to correctly classify misinformation instances and neural ones. In this section, we provide several evaluation criteria for evaluating both the correctness of misinformation detection and timeliness of detection.

\subsection{Correctness of Misinformation Detection} 
Similar to offline misinformation detection, a good system should be able to distinguish between misinformation and neural instances. Following previous work related to binary classification~\cite{DBLP:conf/icde/ZhangDY20,DBLP:conf/acl/PramanickDMSANC21}, we adopt four criteria for evaluating the corectness of detection: Accuracy, Macro-Precision, Macro-Recall and Macro-F1, putting equal importance to both classes.

\subsection{Timeliness of Detection} 
Given a test time misinformation video, as defined in Section~\ref{sec:dataset}, suppose the system predicts $t$ as the start of the misinformation and the misinformation period occurs from $\text{s}^k$ to $\text{e}^k$ ($\text{s}^k$ is the nearest misinformation starting point to $t$). Inspired by previous work about online action detection~\cite{DBLP:conf/eccv/ShouPCMMVNC18,DBLP:conf/iccv/GaoX0SX19}, we adopt point-level precision for evaluation. The misinformation point is considered to be detected in time if $|t-\text{s}^k| \leq \alpha$, where $\alpha$ is a hyper-parameter for the offset tolerance. A discounting factor $\gamma < 1$ will be added when the detection is delayed (detect the $k$-th misinformation span rather than the first one). Optionally, some punishment terms can be added to encourage early detection rather than late detection:
\begin{equation}
    \text{score}_{t} = \gamma^k * \left\{
    \begin{aligned}
         &  1, \quad if  |t-\text{s}^k| \leq \alpha  \ and   \ t \leq \text{s}^k\\
         & \beta, \quad if  |t-\text{s}^k| \leq \alpha  \ and  \ t > \text{s}^k
    \end{aligned}
    \right .
\end{equation} 
where $\beta$ is a hyper-parameter and $\beta<1$. If the system mis-classifies the video as neutral, the score for timeliness will be $0$.

\section{Feasibility Analysis}
\label{sec:analysis}

\subsection{Possible Baselines}
The core idea of baselines is based on~\cite{DBLP:journals/ijcv/NguyenT14}, treating the task into a frame-level binary classification task. We divide baselines into two kinds: 1) uni-modal models, which consider information from one modality or convert all information into one modality; 2) multi-modal models, which consider information from multiple modalities as well as their interactions.

\noindent\textbf{Pre-processing:}
A video may contain information from multiple modalities~\cite{ganti2022novel}, i.e., frame images, texts from subtitles and audio. 
For live streaming videos, we here consider two modalities: 1) visual information from frame images; 2) audio information. For each frame image, we convert it into the image caption $\mathcal{S}_{t}^{\text{I}}$ with any off-the-shelf pre-trained caption generators (i.e., ~\cite{mokady2021clipcap}) and extract the frame-level vector representation $\mathbf{r}_{t}^{\text{I}}$ with either pre-trained image~\cite{he2016deep}  or video representation generators~\cite{qian2021spatiotemporal}. Similarly, we convert the associated audio into texts $\mathcal{S}_{t}^{\text{A}}$ and obtain the audio representation $\mathbf{r}_{t}^{\text{A}}$ with utilization of different pre-trained models.

\noindent\textbf{Uni-modal Models:}
We consider three kinds of unimodal models: 1) receives frame images only; 2) receives audio only; 3) receives the concatenated sentence of $\mathcal{S}_{t}^{\text{I}}$ and $\mathcal{S}_{t}^{\text{A}}$. For the first two cases, the vector representation $\mathbf{o}_t$ for classification is $\mathbf{r}_{t}^{\text{I}}$ and $\mathbf{r}_{t}^{\text{A}}$ respectively. For the last case, we extract the vector representation of the concatenated sentence with language models (i.e., BERT~\cite{devlin2018bert}) as $\mathbf{o}_t$.

\noindent\textbf{Multi-modal Models:}
For multi-modal models, they receive information from both frame images and audio. Different multi-modal fusion techniques, as summarized in~\cite{zhang2019information}, can be applied over $\mathbf{r}_{t}^{\text{I}}$ and $\mathbf{r}_{t}^{\text{A}}$ to generate the joint representation $\mathbf{o}_t$, which serves as the input to the classification layer.

\noindent\textbf{Frame-level Classification:}
Given $\mathbf{o}_{t}$, a classifier is trained to assign higher scores $c_t$ to frames during the early starts of misinformation, while lower scores to negative frames not related to misinformation. The classifier can be either the SOSVM in~\cite{DBLP:journals/ijcv/NguyenT14} or simple MLPs. Some constraints as mentioned in~\cite{DBLP:journals/ijcv/NguyenT14} can be applied and label smoothing (positive frames are not all scored as $1$, while increased from 0 to 1) can also be used. We regard $t$ as the predicted start of misinformation if $c_t$ is above the threshold obtained by grid search on the training set, while $c_{t-1}$ is not. If there is such time step $t$, the video will be regarded as misinformation, otherwise, neutral.

\subsection{Limitation of Baselines}
The baselines do not consider characteristics of the misinformation detection and treat the detection task as an extension from early action recognition. As mentioned in Section~\ref{sec:related}, the detection of misinformation can be through fact checking, propagation or style analysis, which are ignored in the baselines above. On the other hand, baselines omit the temporal relations between frames while treating each frame independently. In some cases, the transaction of frames may also contributes to misinformation (previous frame states vaccine is good while the next frame shows the injection of detergent helps). It calls for more sophisticated models for the detection.

\clearpage

\bibliographystyle{ACM-Reference-Format}
\balance
\bibliography{newref}

\end{document}